\documentclass{article}

\PassOptionsToPackage{numbers, compress}{natbib}

    \usepackage[preprint]{neurips_2019}

\usepackage[utf8]{inputenc} 
\usepackage[T1]{fontenc}    
\usepackage{hyperref}       
\usepackage{url}            
\usepackage{booktabs}       
\usepackage{amsfonts}       
\usepackage{nicefrac}       
\usepackage{microtype}      
\usepackage{graphicx}
\usepackage{fancyvrb}
\usepackage{xcolor}

\title{ViP: Video Platform for PyTorch}

\author{%
  Madan Ravi Ganesh *\\
  University of Michigan \\
  \texttt{madantrg@umich.edu} 
  \And
  Eric Hofesmann \thanks{Denotes equal contribution} \\
  University of Michigan\\
  \texttt{erichof@umich.edu} \\
  \And
  Nathan Louis * \\
  University of Michigan \\
  \texttt{natlouis@umich.edu} \\
  \And
  Jason Corso \\
  University of Michigan \\
  \texttt{jjcorso@umich.edu} 
}

\begin{document}

\maketitle

\begin{abstract} 
This work presents the Video Platform for PyTorch (ViP), a deep learning-based framework designed to handle and extend to any problem domain based on videos. 
ViP supports (1) a single unified interface applicable to all video problem domains, (2) quick prototyping of video models, (3) executing large-batch operations with reduced memory consumption, and (4) easy and reproducible experimental setups.
ViP's core functionality is built with flexibility and modularity in mind to allow for smooth data flow between different parts of the platform and benchmarking against existing methods.
In providing a software platform that supports multiple video-based problem domains, we allow for more cross-pollination of models, ideas and stronger generalization in the video understanding research community. 
\end{abstract}

\section{Introduction} 
\begin{figure}
 \centering
 \includegraphics[width=\columnwidth]{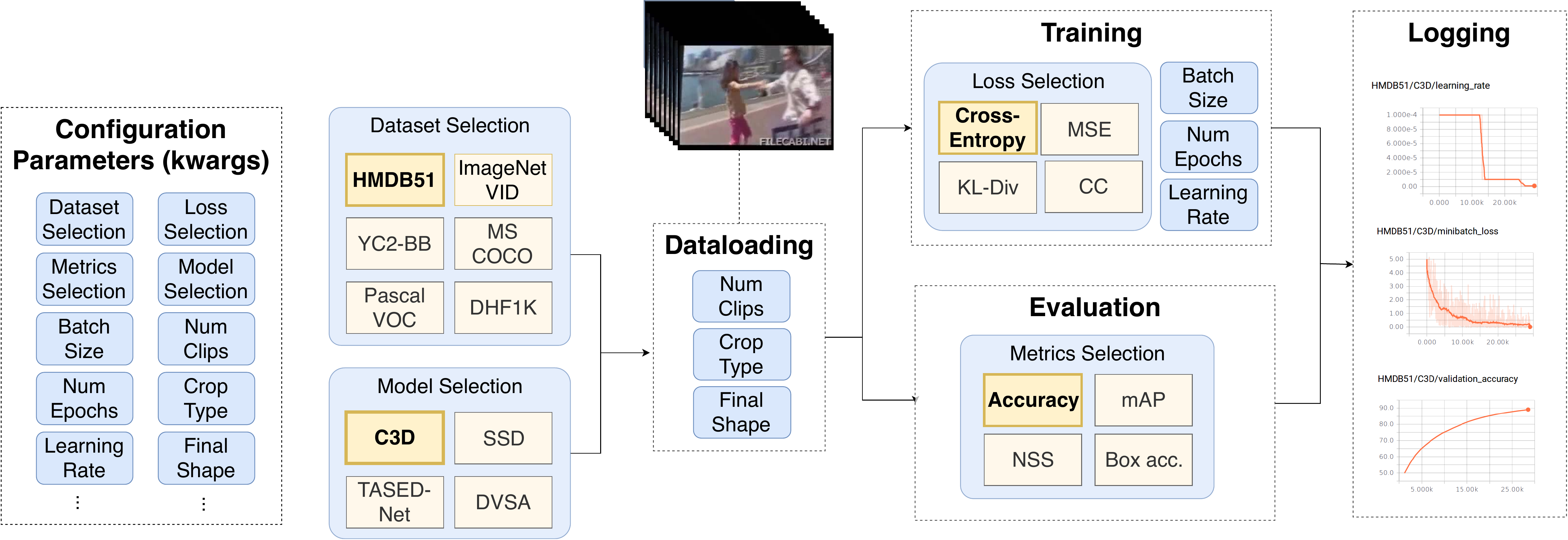}
 \caption{An overview of ViP. 
 All parameters can be specified using the configuration parameters seen on the left.
 Any additional parameters can be added to the configuration file and accessed anywhere within ViP as keyword arguments (\texttt{kwargs}).
 The dataset and model selection are passed through ViP's dataloading which then trains or evaluates using the specified \texttt{kwargs}.
 Output losses, metrics, and other hyperparameters are automatically logged in Tensorboard \cite{abadi2016tensorflow}.}
 \label{fig:main}
\end{figure}

The bulk of recent works in deep learning have focused on tasks involving videos, be it in applications like reinforcement learning~\cite{babaeizadeh2016reinforcement, andrychowicz2017hindsight} and surveillance~\cite{nascimento2006performance,mabrouk2018abnormal} or in conventional image oriented tasks such as captioning~\cite{yang2018video,zhou2018end,pan2017video} and detection~\cite{zhu2018towards, zhu2017flow}.
Although the specifics of the loss and datasets used are unique to each task, the general principles used to design models and the way videos are handled remain consistent across most problem domains.
In an effort to lead the development of models with strong generalization properties we introduce ViP, the Video Platform for PyTorch.
The platform presents itself as a unique collection of video-specific processing methods that interact fluidly with components of a standard experimental pipeline.
Each of these components were designed as stand-alone functions to ensure cross-compatibility with any dataset, model, loss, or metric chosen by the user.
Figure \ref{fig:main} gives a high level overview of how these components interact with each other.
This design ensures that a variety of combinations can be quickly setup and trained, while ensuring comparability with standard benchmarks.

The main contributions presented in ViP are,
\begin{itemize}
    \item A common end-user interface for all supported problem domains
    \item Removal of overheads in prototyping new models and setting up experiments
    \item Support for large-batching operations on systems with limited memory, affording smaller nodes a chance to compete against large-scale professional clusters
    \item Customized internal logging and bookkeeping allowing for continuation of experiments from a desired point in time as well as repeatability and reproducibility
\end{itemize}
Our code is made publicly available\footnote{\url{https://github.com/MichiganCOG/ViP}}.

\section{Related Works} 

Other platforms have provided code bases with the intention of deploying and benchmarking new or existing models. However these frameworks are generally task specific, targeting at most one or two problem domains.
Detectron \cite{Detectron2018} and MMDetection \cite{mmdetection} each provide a platform for object detection research. However, neither is inherently extendible to videos or video object detection methods. The recently released Pythia \cite{singh2018pythia} is built upon vision and language problems, but is primarily used for image datasets. M-PACT \cite{hofesmann2018m} and MMAction \cite{mmaction2019} are both applied to video data, but only for action recognition or activity classification.

All of the aforementioned systems support only their tailored vision task. In contrast, ViP is generalizable to multitudes of computer vision problems involving either image or video.
We currently support activity recognition, object detection, visual saliency prediction, and video object grounding as detailed in Sec. \ref{sec:datasets}. Our aim is to provide a scalable platform for a breadth of problems in computer vision that supports new and exiciting research.

\section{Breakdown of ViP}

\subsection{Dataloading} \label{sec:dataloading} 
ViP offers unique options to load data, specifically videos, that are not common in other platforms.
These options include a flexible way of extracting clips from videos, video-based preprocessing functions, bounding box and keypoint preprocessing, and a consistent JSON format to automatically load datasets.

\begin{figure}
    \centering
    \includegraphics[width=0.75\textwidth]{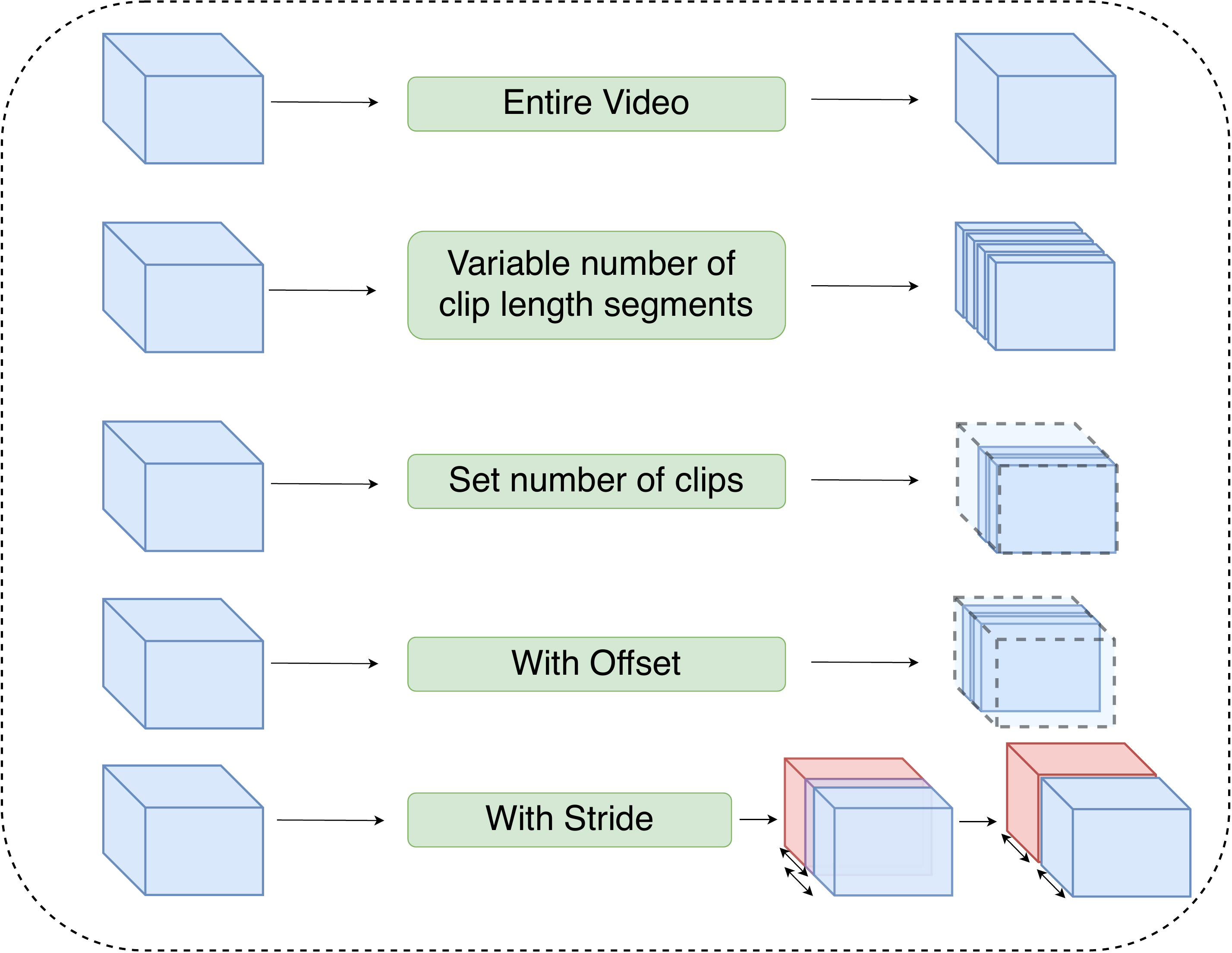}
    \caption{Most networks require input videos to be of a fixed length. To facilitate this, videos are often broken down into clips.
    In ViP, there are a set of arguments that can be used to determine how clips are extracted from a given video.}
    \label{fig:extract_clips}
\end{figure}

\paragraph{Clip Extraction} 
Extracting clips from videos is vital to ensure that the inputs to a network are of a uniform shape since most datasets contain variable length videos.
There exists no standard method to extract clips, it is unique to each model and experimental setup.
ViP contains a set of parameters that describe how to extract clips from videos in a wide variety of ways, as seen in Fig.~\ref{fig:extract_clips}.
Examples of these parameters include \texttt{clip\_length}, \texttt{num\_clips}, \texttt{clip\_stride}, and \texttt{clip\_offset}.
These parameters can be used to uniformly sample frames to get one clip per video, extract sequential clips up to the length of a video, randomly sample a clip from somewhere in the video, and many more.

\paragraph{Video-based Preprocessing}
PyTorch provides a set of image transforms based on the Python Imaging Library (PIL) which can be used to augment images before they are passed into a neural network.
While ViP is compatible with these transforms, problems arise when using the image-based methods on videos.
For example, randomly cropping every frame of a video will result in a clip where every sequential frame jumps from one cropped portion of the image to another.
In ViP, methods such as cropping, resizing, flipping, rotating, and mean value subtraction are implemented to be constant across an entire video.
Additionally, ViP contains methods \texttt{ApplyToPIL} and \texttt{ApplyOpenCV} which can take any PIL or OpenCV~\cite{opencv_library} function and apply it to an entire clip.

\paragraph{Bounding Box Augmentation}
When preprocessing videos which contain bounding box annotations, any augmentation that is used on the video frames must also be applied to the bounding box coordinates.
All of the preprocessing methods implemented in ViP are constructed so that if given both a video clip and the corresponding bounding box (or keypoint) annotations, the same transform will be applied to both.
For example, if a video clip is cropped randomly then the bounding boxes of objects in the video will be adjusted according to the crop location. 

\paragraph{JSON Dataset Format}
To facilitate the loading of a dataset into the training or evaluation script of ViP, we have developed a consistent JSON format that must be generated for a given dataset.
After conversion to the JSON format, a dataset can be loaded and the aforementioned functions can be used with no additional work by the user.
These JSON files are composed of a list of dictionaries which contain the ground truth annotations and paths to the videos in a dataset.
All datasets in ViP, described in Sec.~\ref{sec:datasets}, contain a script to generate this JSON file automatically.
This JSON format is robust enough to allow for the addition of any video dataset no matter the type of annotations.
For example, it currently supports action classes, bounding boxes, keypoints, and saliency maps.

{}


\subsection{Executing experiments} 
Experiments across all problem domains use the same training and evaluation functions. We define the experiment's parameters as system arguments in a YAML configuration file (Fig. 3). These parameters define the dataset, model, loss, metric, preprocessing functions (from Sec. \ref{sec:dataloading}), and many other options. Note that many of these can be overridden with command line arguments. 


\begin{minipage}{.48\textwidth}
\raggedleft
\begin{Verbatim}[commandchars=\\\{\}]
  \textcolor{gray}{1 \quad # Preprocessing}
  \textcolor{gray}{2}  \quad \textcolor{teal}{clip_length:}  \quad \textcolor{blue}{16}
  \textcolor{gray}{3}  \quad \textcolor{teal}{clip_offset:}  \quad \textcolor{blue}{0}
  \textcolor{gray}{4}  \quad \textcolor{teal}{clip_stride:}  \quad \textcolor{blue}{0}
  \textcolor{gray}{5}  \quad \textcolor{teal}{crop_shape:}   \quad \textcolor{blue}{[112,112]}
  \textcolor{gray}{6}  \quad \textcolor{teal}{crop_type:}    \quad \textcolor{blue}{Random}
  \textcolor{gray}{7}  \quad \textcolor{teal}{final_shape:}  \quad \textcolor{blue}{[112,112]}
  \textcolor{gray}{8}  \quad \textcolor{teal}{num_clips:}    \quad \textcolor{blue}{-1}
  \textcolor{gray}{9}  \quad \textcolor{teal}{random_offset:}\quad \textcolor{blue}{0}
  \textcolor{gray}{10} \quad \textcolor{teal}{resize_shape:} \quad \textcolor{blue}{[128,171]}
  \textcolor{gray}{11} \quad \textcolor{teal}{subtract_mean:}\quad \textcolor{blue}{`'}
  \textcolor{gray}{12} 
  \textcolor{gray}{13}
  \textcolor{gray}{14 \quad # Experimental Setup}
  \textcolor{gray}{15} \quad \textcolor{teal}{acc_metric:}   \quad \textcolor{blue}{Accuracy}
  \textcolor{gray}{16} \quad \textcolor{teal}{batch_size:}   \quad \textcolor{blue}{3}
  \textcolor{gray}{17} \quad \textcolor{teal}{dataset:}      \quad \textcolor{blue}{HMDB51}
  \textcolor{gray}{18} \quad \textcolor{teal}{debug:}        \quad \textcolor{blue}{0}
  \textcolor{gray}{19} \quad \textcolor{teal}{epoch:}        \quad \textcolor{blue}{30}
\end{Verbatim}
\end{minipage}
\begin{minipage}{0.48\textwidth}
\begin{Verbatim}[commandchars=\\\{\}]
 \textcolor{gray}{20} \quad \textcolor{teal}{exp:}              \quad \textcolor{blue}{exp}
 \textcolor{gray}{21} \quad \textcolor{teal}{gamma:}            \quad \textcolor{blue}{0.1}
 \textcolor{gray}{22} \quad \textcolor{teal}{grad_max_norm:}    \quad \textcolor{blue}{10}
 \textcolor{gray}{23} \quad \textcolor{teal}{json_path:}        \quad \textcolor{blue}{/path/HMDB51}
 \textcolor{gray}{24} \quad \textcolor{teal}{labels:}           \quad \textcolor{blue}{51}
 \textcolor{gray}{25} \quad \textcolor{teal}{load_type:}        \quad \textcolor{blue}{train}
 \textcolor{gray}{26} \quad \textcolor{teal}{loss_type:}        \quad \textcolor{blue}{M_XENTROPY}
 \textcolor{gray}{27} \quad \textcolor{teal}{lr:}               \quad \textcolor{blue}{0.0001}
 \textcolor{gray}{28} \quad \textcolor{teal}{milestones:}       \quad \textcolor{blue}{[10,20]}
 \textcolor{gray}{29} \quad \textcolor{teal}{model:}            \quad \textcolor{blue}{C3D}
 \textcolor{gray}{30} \quad \textcolor{teal}{momentum}          \quad \textcolor{blue}{0.9}
 \textcolor{gray}{31} \quad \textcolor{teal}{num_workers:}      \quad \textcolor{blue}{2}
 \textcolor{gray}{32} \quad \textcolor{teal}{opt:}              \quad \textcolor{blue}{sgd}
 \textcolor{gray}{33} \quad \textcolor{teal}{preprocess:}       \quad \textcolor{blue}{default}
 \textcolor{gray}{34} \quad \textcolor{teal}{pretrained:}       \quad \textcolor{blue}{1}
 \textcolor{gray}{35} \quad \textcolor{teal}{pseudo_batch_loop:}\quad \textcolor{blue}{1}
 \textcolor{gray}{36} \quad \textcolor{teal}{rerun:}            \quad \textcolor{blue}{1}
 \textcolor{gray}{37} \quad \textcolor{teal}{save_dir:}         \quad \textcolor{blue}{`./results'}
 \textcolor{gray}{38} \quad \textcolor{teal}{seed:}             \quad \textcolor{blue}{999}
 \textcolor{gray}{39} \quad \textcolor{teal}{weight_decay:}     \quad \textcolor{blue}{0.0005}
\end{Verbatim}
\end{minipage} \\
\quad \\
Figure 3: Example of a YAML configuration file specifying the experiment's parameters.
Any desired parameters can be added to this file and accessed through \texttt{kwargs} anywhere within ViP.
For example, if a newly defined model or loss requires a unique variable, it can be specified in this file.

\subsubsection{Training} 
Training a new model requires only the development of a model architecture file, and the specification of a dataset and loss. As part of dataloading in ViP, we define preprocessing functions to be model specific rather than dataset specific. This retains the modularity of this framework, swapping out different datasets and losses between experiments, without writing any new code. As part of ViP, novel or multi-part losses can be easily added and later referenced in the configuration file.

\subsubsection{Evaluation and Metrics} 
The only requirements for evaluation are the model architecture file, a dataset, and a metric. These are all specified in the YAML configuration file. Currently the  standard metrics provided in ViP are: Accuracy, Intersection-over-union (IoU), Average Precision (AP), mean Average Precision (mAP), Normalized Scanpath Saliency (NSS), and Linear Correlation Coefficient (CC). Additionally, any of these metrics can be inherited from to create unique metrics tailored to a model's output. We plan to include other standard metrics as we expand across problem domains. 


\subsection{Implementation Quirks} 
Throughout the previous sections we described the general process flow in ViP and how a suite of related metrics and other parts help provide all of the general functionality required from a platform.
In this section we highlight three specific implementation quirks which elevate the general functionality of our platform to perform tasks that are not possible elsewhere.

     \paragraph{Keyword Arguments} 
     We co-opt \texttt{kwargs}, available in python, to pass arguments throughout the entire platform. There is no restriction on the number of arguments that can be passed to (or from) any function and this can be done without changing any function prototype/header.
    \paragraph{Pseudo Batches}
    In order to encourage competition against the large-scale compute capabilities available in professional environments we include pseudo-batching~\cite{morton_2018}. It is a form of large scale batching that accumulates gradients across multiple mini-batches. Thus, even in single-GPU systems, ViP can perform large-batch computations with minimal memory requirements. By extension, this allows for pseudo-batches with variable sized inputs, a functionality that, to our knowledge, is unavailable on any other platform.
    \paragraph{Loading Weights} 
    The \texttt{pretrained} keyword carries special significance in ViP. It allows for 3 distinct values, 0, 1 and the full path to a trained checkpoint from ViP. The overloaded keyword can signify  random initialization (0), loading pretrained weights (1), or a checkpoint from a model trained in ViP (path). As an extension of this functionality the end-user can choose to pickup training where the checkpoint left off or alter the objective function completely.


\subsection{Datasets and Benchmarks} 
\label{sec:datasets}
As mentioned in Sec. \ref{sec:dataloading}, ViP requires datasets to have a consistent JSON format to be loaded into the framework. This requires a simple restructure and compilation of the annotation files. Currently, we've provided scripts to generate these JSON files for the following datasets: HMDB51~\cite{kuehne2011hmdb}, UCF101~\cite{soomro2012ucf101}, Kinetics-400~\cite{carreira2017quo}, ImageNetVID 2015~\cite{ILSVRC15}, DHF1K~\cite{dhf1k}, MSCOCO 2014~\cite{lin2014microsoft}, and Pascal VOC 2007~\cite{pascal-voc-2007}. In ViP we have also implemented the following models: C3D~\cite{tran2015learning}, SSD300~\cite{liu2016ssd}, TASED-Net~\cite{min2019tased}, DVSA (with frame-wise weighting and object interaction)~\cite{zhou2018weakly}. Results for implemented benchmarks are shown in Table \ref{tab:results}.

\paragraph{Action Recognition}
One of the earliest and most widely studied tasks in video literature is action recognition.
Currently, in ViP we support HMDB51, UCF101 and Kinetics-400 directly while giving the end-users the ability to include custom datasets.
HMDB51 is one of the earliest datasets to span a diverse range of actions from multiple sources. 
It contains 3570 videos for training and 1530 for testing, across 51 labels.
UCF101 succeeded this dataset by expanding the total number of labels to 101 and almost tripling the total number of training and testing videos to 9537 and 3783 respectively.
More recently, Kinetics-400 expanded the number of diverse actions to 400 across \textasciitilde300K videos.

\paragraph{Object Detection}
Traditionally, object detection refers to \textit{image} object detection which is the task of localizing an object, typically with a bounding box, from a known list of classes.
In ViP, we support datasets for both images and videos. Pascal VOC 2007 \cite{pascal-voc-2007} has a total of 9963 images with 20 object categories. MSCOCO 2014 \cite{lin2014microsoft} has a total of 328,000 images with 91 object categories. ImageNetVID 2015 \cite{ILSVRC15} contains 3862, 555, and 937 training, validation, and testing videos (respectively) across 30 object classes.

\paragraph{Video Saliency} The goal of video saliency prediction models is to predict the gaze location of a human watching a video.
ViP supports the video saliency dataset DHF1K~\cite{dhf1k}, which contains a ground-truth map of binary pixel-wise gaze fixation points and a continuous map of the fixation points after being blurred by a gaussian filter.
DHF1K contains 1000 videos in total.
700 of the videos are annotated, 600 of which are used for training and 100 for validation. 
The remaining 300 are the testing set which are to be evaluated on a public server.

\paragraph{Video Object Grounding}
Video object grounding is the task of aligning words in a given sentence to relevant objects or target regions in a video. A word from the sentence that is correctly localized, with a bounding box, is said to be ``grounded''. YouCook2-BoundingBox \cite{zhou2018weakly} provides videos used for the video object grounding task. It includes a total of 4325 annotated segments from 67 grounded words.

%
%
%

\begin{table}[h]
   \caption{Results of the models provided in ViP across all currently implemented tasks. (fw - framewise weighting, obj - object interaction)
   }
   \label{tab:results}
   \centering
\begin{tabular}{cccl}
\toprule
\textbf{Model Architecture} & \textbf{Task} & \textbf{Dataset} & \textbf{ViP Metric (\%)}\\
\midrule 
C3D    & Activity Recognition & HMDB51 (Split 1) & $50.14$ (acc.) \\
C3D    & Activity Recognition & UCF101 (Split 1) & $80.40$ (acc.) \\
SSD300 & Object Detection     & VOC2007          & $76.58$ (AP) \\
TASED-Net & Video Saliency & DHF1K (Validation) & $51.50$ (CC) \\
DVSA (+ fw,obj)  & Video Object Grounding & YC2-BB (Validation) & $30.09$ (box acc.)\\
\bottomrule
\end{tabular}
\end{table}

\section{Conclusion} 

We have developed ViP, the video platform for deep learning in PyTorch.
While other platforms are built around images or specific video tasks, ViP offers a flexible framework for a wide range of tasks that abstracts away time consuming implementation details. 
Additionally, ViP provides a variety of features to expedite neural network development, training, evaluation, and benchmarking.
In the future, more video-based models and datasets will be added across an even wider range of tasks.

{\small
\bibliographystyle{abbrv}
\bibliography{egbib}
}







\end{document}